\documentclass[runningheads]{llncs}
\usepackage{graphicx}
\usepackage{times}
\usepackage{epsfig}
\usepackage{graphicx}
\usepackage{amsmath}
\usepackage{amssymb}
\usepackage{booktabs}
\usepackage{caption}
\usepackage{multirow, multicol}
\usepackage{color}
\usepackage{hyperref}       
\usepackage{url} 
\usepackage{booktabs}       
\usepackage{amsfonts}       
\usepackage{nicefrac}       
\usepackage{microtype}      
\usepackage{xcolor} 
\usepackage{graphicx}
\usepackage{multirow}
\usepackage{amsmath}
\usepackage{subfigure}
\usepackage{wrapfig}
\usepackage{marvosym}

\definecolor{up}{RGB}{68,169,32}
\definecolor{down}{RGB}{255,0,0}
\hypersetup{
colorlinks=true,
linkcolor=blue,
citecolor=blue
}

\begin{document}
%
\title{Diffusion-based 3D Object Detection with Random Boxes}

\author{Xin Zhou\inst{1} \and
Jinghua Hou\inst{1} \and
Tingting Yao\inst{1} \and
Dingkang Liang\inst{1} \and
Zhe Liu\inst{1} \and
Zhikang Zou\inst{2} \and
Xiaoqing Ye\inst{2} \and
Jianwei Cheng\inst{3} \and
Xiang Bai\inst{1}\textsuperscript{(\Letter)}
}
\authorrunning{X. Zhou et al.}

\institute{Huazhong University of Science and Technology, Wuhan, China \email{\{xzhou03,xbai\}@hust.edu.cn}
\and
Baidu Inc., China \and JIMU Intelligent Technology Co., Ltd, China}
\maketitle              
\begin{abstract}
3D object detection is an essential task for achieving autonomous driving. Existing anchor-based detection methods rely on empirical heuristics setting of anchors, which makes the algorithms lack elegance. In recent years, we have witnessed the rise of several generative models, among which diffusion models show great potential for learning the transformation of two distributions. Our proposed Diff3Det migrates the diffusion model to proposal generation for 3D object detection by considering the detection boxes as generative targets. During training, the object boxes diffuse from the ground truth boxes to the Gaussian distribution, and the decoder learns to reverse this noise process. In the inference stage, the model progressively refines a set of random boxes to the prediction results. We provide detailed experiments on the KITTI benchmark and achieve promising performance compared to classical anchor-based 3D detection methods.

\end{abstract}
\section{Introduction}
\begin{figure}[t]
	\begin{center}
		\includegraphics[width=0.92\linewidth,height=0.23\linewidth]{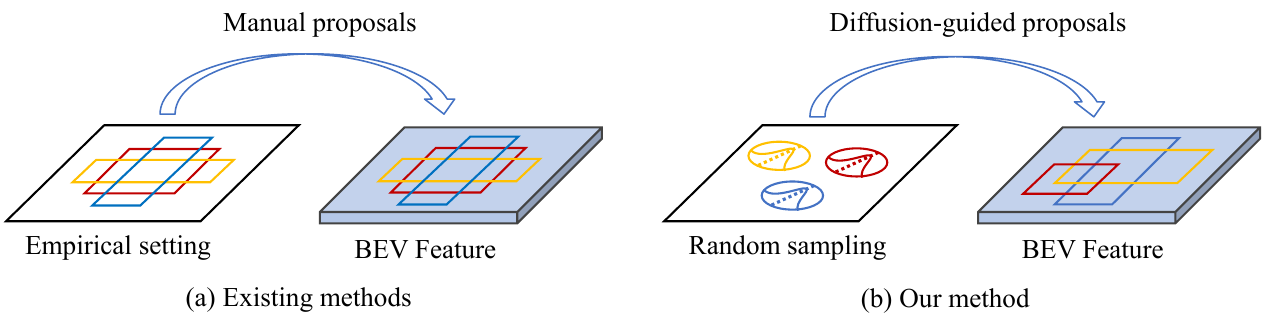}
	\end{center}
        \vspace{-13pt}
	\caption{Compared with existing anchor-based 3D object detection paradigms. (a) Manual set proposals; (b) Diffusion-guided proposals (ours). The existing methods rely on manual anchors for prediction, while ours requires only Gaussian noises.}
 \vspace{-10pt}
	\label{fig: intro }
\end{figure}
3D object detection, a fundamental task in computer vision, aims to regress the 3D bounding boxes and recognize the corresponding category from the point clouds. It is widely used in autonomous driving as a core component for 3D scene understanding. However, due to the intrinsic sparsity and irregularity of the derived point clouds, building high-accuracy LiDAR-based 3D object detection methods is challenging.

Recently, the mainstream approaches can be divided into two categories according to the representation formats of the point clouds: point-based~\cite{shi2019pointrcnn,zhang2022not} and voxel-based methods~\cite{zhou2018voxelnet,yan2018second,lang2019pointpillars,liu2020tanet,shi2020points,chen2023voxelnext}. Although point-based methods have achieved reliable performance on object localization, they are challenging to handle large-scale point clouds scenarios due to high computation costs of point clouds sampling and grounding operations~\cite{qi2017pointnet++}. In contrast, the voxel-based manners can convert irregular raw point clouds into regular voxel grid format and implement efficient feature extraction on voxels by highly optimized 3D sparse convolution operations. Thus, to trade off the performance and efficiency, the network in this paper is mainly based on the voxel representation. However, existing voxel-based methods~\cite{yan2018second,zhou2018voxelnet,lang2019pointpillars,liu2020tanet,yin2021center} still rely on empirical heuristics to set anchor sizes or center radii, which might not be an elegant strategy, as shown in Fig.~\ref{fig: intro }(a). This leads us to ask \textit{whether it is feasible to recover predictive boxes from more succinct random boxes directly}.

Fortunately, we have witnessed the rise of diffusion models~\cite{ho2020denoising,song2020denoising}, a probabilistic modeling paradigm that demonstrates its potential in many 2D vision tasks (e.g., image generation~\cite{liu2022compositional,rombach2022high,gu2022vector}, segmentation~\cite{chen2022generalist,amit2021segdiff,brempong2022denoising} and 2D object detection~\cite{chen2022diffusiondet}). Among these, DDPM~\cite{ho2020denoising} stands out as a seminal advancement. It treats the process of image generation as a Markov process, wherein Gaussian noise is intentionally introduced to the target distribution. Through the training of a dedicated network, this noise is subsequently removed, facilitating the restoration of the pristine underlying structure. By learning the intricate mapping from a Gaussian distribution to the target distribution, diffusion models inherently possess the intrinsic capability to denoise ideal data. Despite their resounding success in 2D vision tasks, the untapped potential of diffusion models in proposal generation for 3D object detection remains unexplored.

In this paper, we present a framework named Diff3Det to explore the feasibility of generative models for 3D object detection. Specifically, the model adds Gaussian noise with a controlled variance schedule to the ground truth boxes in the training stage to obtain noisy boxes, as shown in Fig.~\ref{fig: intro }(b). These noisy boxes are then used to extract Region of Interest (RoI) features from the BEV feature map, which does not need to set the manual anchor. The detection decoder then incorporates these features and time planes to predict offsets between the noisy and ground truth boxes. As a result, the model can recover the ground truth boxes from the noisy ones. We reverse the learned diffusion process during inference to generate bounding boxes that fit a noisy prior distribution to the learned distribution over the bounding boxes.

Our main contributions of this paper can be summarized in two folds as follows:
\begin{itemize}
\item We present a framework named Diff3Det that explores the feasibility of generative models for 3D object detection, achieving promising performance compared with the popular 3D object detectors, and demonstrating the potential of diffusion models in 3D vision tasks.
\item We design several simple strategies to select proposal boxes to address the sparsity of point cloud data and 3D features to improve the usability of the methods. In addition, we propose an optimized noise variance scheduling for diffusion models that can be better adapted to the 3D object detection task.
\end{itemize}

\section{Related Work}
\subsection{3D Object Detection}

Most 3D object detection methods rely on LiDAR. LiDAR-based detection has two main streams: point-based and voxel-based. Point-based methods~\cite{shi2019pointrcnn,yang20203dssd,zhang2022not,li2023dds3d,zhang2023sam3d,2023asimple} directly learn geometry from unstructured point clouds and generate object proposals. However, these methods often have insufficient learning capacity and limited efficiency. In contrast, VoxelNet~\cite{zhou2018voxelnet} converts the irregular point clouds into regular voxel grids. To improve computational efficiency, some methods~\cite{yan2018second,liu2020tanet,deng2021voxel,sparsedet,chen2023voxelnext} leverage highly optimized 3D submanifold sparse convolutional networks. Due to the assumption of no repeating objects in height in autonomous driving, some methods~\cite{lang2019pointpillars,shi2022pillarnet} only voxelize in the plane and apply 2D convolution to further improve computational efficiency. Although voxel-based methods are computationally efficient and better suited for feature extraction, they inevitably introduce information loss. Researchers~\cite{shi2020pv,noh2021hvpr} utilize both point-based and voxel-based representations for further learning. 

Different from LiDAR-based 3D object detection, image-based methods can significantly reduce sensor costs. Image-based methods~\cite{liu2022petr,li2023bevdepth,xiong2023cape} for 3D object detection estimate depth and detect objects from 2D images, but the performance of these methods is still limited. To overcome this challenge, multimodal-based 3D object detection methods~\cite{huang2020epnet,bai2022transfusion,liu2022epnet++,li2023logonet} combine precise geometric information from LiDAR with rich semantic information from images, resulting in state-of-the-art performance.

Here, we notice that many previous approaches~\cite{zhou2018voxelnet,yan2018second,lang2019pointpillars,liu2020tanet} still require manual selection of anchor boxes in advance for subsequent proposal generation, which largely depends on the human experience. To the best of our knowledge, more elegant ways to generate proposals are still under-explored.

\subsection{Diffusion Models in Vision}
Diffusion is a physical model aimed at minimizing the spatial concentration difference. In the computer vision field, Diffusion~\cite{ho2020denoising,song2020denoising} is a probabilistic model that uses a forward process to transform the initial distribution into a normal distribution and train a network to reverse the noise. The diffusion model has shown promising results in many tasks, including image generation~\cite{liu2022compositional,rombach2022high,gu2022vector}, segmentation~\cite{amit2021segdiff,chen2022generalist,brempong2022denoising} and depth estimation~\cite{ji2023ddp,duan2023diffusiondepth}. Recently, DiffusionDet~\cite{chen2022diffusiondet} extends the diffusion process into generating detection box proposals, showing that the prospects for applying diffusion models to detection tasks are bright. Inspired by DiffusionDet, we explore the application of Diffusion models on proposals generation in 3D object detection. 

\section{Proposed Method}
In this section, we first revisit the diffusion model in Sec.~\ref{revisit}. Then, we introduce the overall of our method (Sec.~\ref{overall}) and details of the proposed proposal generator (Sec.~\ref{generator}) as well as training and inference processes separately.

\subsection{A Revisit of Diffusion Models} \label{revisit}

The diffusion model~\cite{ho2020denoising,song2020denoising} is a powerful generative model that generates high-quality samples from Gaussian noise, whose pipeline consists of forward and backward processes. Specifically, it builds the forward process by gradually adding noise to a sample and transforming the sample into a latent space with an increasing noise, which follows the Markov chain. The forward process is formulated as follows:

\begin{equation}
q\left(x_{t} \mid x_{0}\right)=\mathcal{N}\left(x_{t} ; \sqrt{\bar{\alpha}_{t}} x_{0},\left(1-\bar{\alpha}_{t} \right) I\right),
\end{equation}
\begin{equation}
\bar{\alpha}_{t} := {\textstyle \prod_{i=0}^{t}} \alpha_{i}={\textstyle \prod_{i=0}^{t}}(1-\beta_{i}),
\end{equation}
where $x_{0}$, $x_{t}$, and $\beta_{i}$ represent the sample, latent noisy sample, and noise variance schedule, respectively. During training, a neural network $f_{\theta}(x_{t},t)$ is trained to predict the original sample $x_{0}$ from the noisy sample $x_{t}$ at each time step $t$ by minimizing the $\ell_{2}$ loss between the predicted and original sample.
\begin{equation}
\mathcal{L}_{train}=\frac{1}{2}\left \|  f_{\theta}(x_{t},t)-x_{0} \right \|^{2} .
\end{equation}
The model reconstructs the original data sample from the noisy sample at the inference stage by iteratively applying the updating rule in reverse order.

In our work, we attempt to apply the diffusion model to the 3D object detection task. We consider the ground-truth bounding boxes as $x_{0}$, where $x_{0}\in\mathbb{R} ^{N\times5}$. A network $f_{\theta}(x_{t},t,x)$ is trained to predict $x_{0}$ from noisy boxes $x_{t}$ by the corresponding point clouds features $x$.

\subsection{Overall} \label{overall}
\begin{figure}[t]
	\begin{center}
		\includegraphics[width=1.0\linewidth]{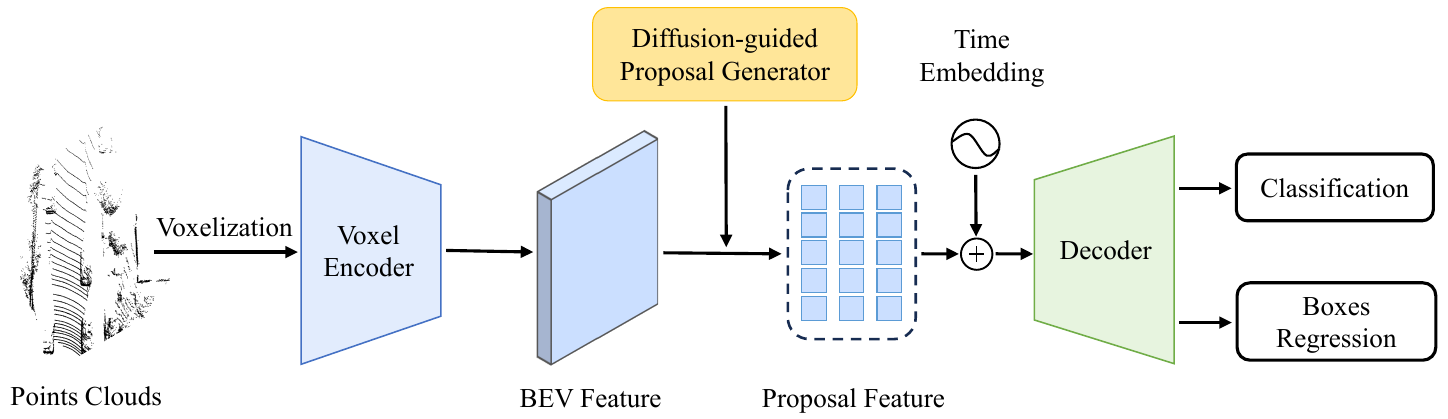}
	\end{center}
        \vspace{-10pt}
	\caption{The pipeline of our method. The point clouds are fed to a 3D encoder for generating BEV features. Then, the diffusion-guided proposal generator generates some random proposals on the BEV. Finally, the detection decoder consumes BEV proposal features and time embeddings to predict the detection results.}
	\label{fig:pipeline }
\end{figure}

Our Diff3Det consists of a diffusion-guided proposal generator, an encoder, and a decoder, as shown in Fig.~\ref{fig:pipeline }. The diffusion-guided proposal generator generates corrupted boxes $x_{t}$ by adding Gaussian noise on the ground truth boxes. The encoder, a 3D voxel backbone~\cite{yan2018second}, is utilized to extract the features of point clouds. The decoder aims to predict the original ground truth boxes by the corrupted boxes $x_{t}$ and corresponding region of interest (RoI) features. Specifically, we utilize the dynamic head~\cite{sun2021sparse}, extended to 3D object detection. Our approach does not rely on the learnable query and learnable embedding for dynamic convolution prediction but instead adopts randomly selected proposal boxes, temporal noise levels, and RoI features.

\subsection{Diffusion-guided Proposal Generator} \label{generator}
\begin{figure}[t]
	\begin{center}
		\includegraphics[width=0.95\linewidth]{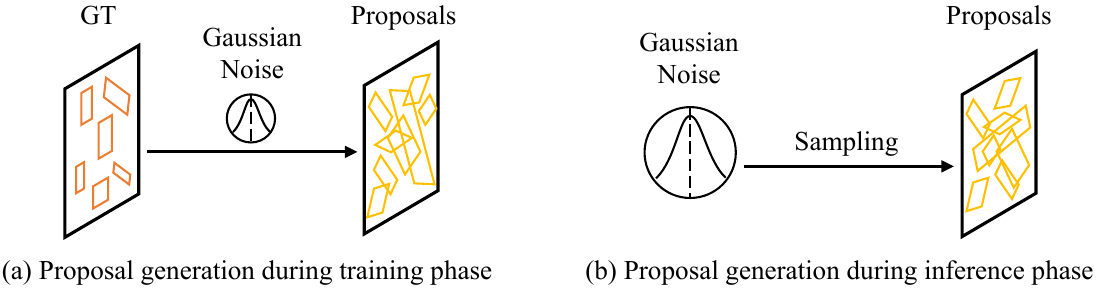}
	\end{center}
       \vspace{-10pt}
	\caption{Diffusion-guided Proposal Generator. Our diffusion-guided proposal generator generates proposals in the training (a) and inference (b) phases by adding Gaussian noise on the ground truth and sampling from Gaussian noise.}
	\label{fig:generater}
\end{figure}
Bird's eye view (BEV) is an effective representation for 3D object detection. Therefore, our method uses the BEV boxes $\left ( cx, cy, dx, dy,\theta \right )$ for the diffusion process. For constructing the initial boxes $x_{0}$, we repeat the ground truth boxes to $N$ and normalize them between 0 and 1. A signal scaling factor controls the diffusion process's signal-to-noise ratio (SNR)~\cite{chen2022analog}. Then, we generate the corrupted boxes $x_{t}$ by adding Gaussian noise to $x_{0}$, which is formulated as:
\vspace{-5pt}
\begin{equation}
x_{t}=\sqrt{\bar{\alpha}_{t}} x_{0}+\sqrt{1-\bar{\alpha}_{t}} \varepsilon ,
\end{equation}
where $\varepsilon\sim \mathcal{N}\left(0,I_{5} \right)$, $t=\mathrm{randint} (1, T_{\mathrm{max}})$, $\bar{\alpha}_{t}$ is the same as~\cite{song2020denoising}. The maximum time $T_{\mathrm{max}}$ is set to an integer (e.g., 1000).

As shown in Fig.~\ref{fig:generater}(a), the diffusion-guided proposal generator generates proposal boxes from the ground truth during training. Firstly, a proposal box with no point makes it tough to recover the target. We adopt a resampling operation by calculating the number of points $m$ in each proposal box. If $m<\eta$, remove the boxes and resample random boxes. Repeat this loop until every proposal box with at least $\eta$ points in it. Moreover, we find that the quality of the proposal boxes is the key to the success of our method, so we adopt simple ways to refine the proposal boxes, which will be explained in detail in the following sections.

\subsubsection{Correlation Coefficient on Size.} 
\begin{figure}[t]
	\begin{center}
		\includegraphics[width=1.0\linewidth]{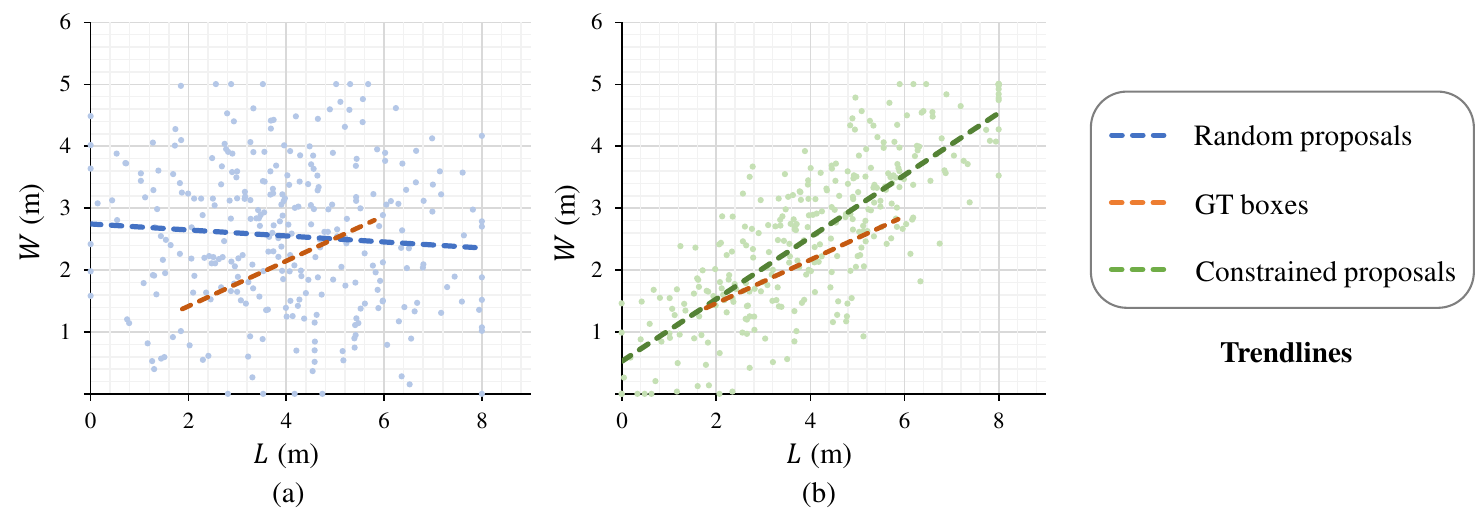}
	\end{center}
    \vspace{-13pt}
	\caption{Distribution of Random proposals vs. Constrained proposals. The distribution of our constrained proposals is more correlated with GT than random proposals.}
    \vspace{-1em}
	\label{fig: correlation }
\end{figure}
It is clear that in the real world, there is a definite relationship between the width and length of a 3D detection box. However, the two independent random distributions of width and length will produce unrealistic proposals, as shown in Fig.~\ref{fig: correlation }(a). For this reason, it is inappropriate to consider the size $\left ( w,l\right )$ as two independent and identically distributed random variables. Therefore, we introduce a correlation coefficient to restrict the box size for those resampled boxes and the boxes used during inference. 
\begin{equation}
W = \rho L + \sqrt{1-\rho ^ {2}}X,
\end{equation}
where $L,X \overset{\mathrm{i.i.d.}}{\sim} \mathcal{N}(0,1)$, and we set the value of correlation coefficient $\rho=0.8$. After generating the random vector $(W, L)$, we scale them to the ranges $(0,w)$ and $(0,l)$ as the width and length of the proposals. We set $w = 8, l = 5$ to satisfy the target. As shown in Fig.~\ref{fig: correlation }(b), after the correlation coefficient constraint, the distribution of generated proposals more correlated with ground truth.

\subsubsection{Dynamic Time Step.}

In the early training phase, recovering samples from seriously corrupted samples with high noise levels is difficult, which harms the final performance. Therefore, we propose a sine schedule to control the time step range, where the noise gradually increases in training. Specifically, $n$ is the total training epoch number, $x$ is the current epoch index, and $T$ is the maximum time to reach. The maximum time on one training epoch $T_{max}$ can be calculated as:
\begin{equation}
 T_{max}=\left\{
\begin{array}{rcl}
&T \left \lfloor\sin \left ( \frac{\cos^{-1} \left ( \frac{\omega }{T}  \right ) }{\sigma  n} x +\sin^{-1}\left (  \frac{\omega }{T}\right )  \right ) \right \rfloor  , {x < \sigma  n}
\vspace{5pt}
\\ & \quad \quad \quad \quad \quad \quad T \quad \quad \quad \quad \quad \quad \quad \quad ~ , {x \ge \sigma  n}
\end{array} \right.
\end{equation} 
where $\omega$ and $\sigma$ are hyperparameters that control the initial time steps at the first epoch and control the training time-point to reach the maximum time $T$, respectively. We empirically set $\omega = 5$ and $\sigma = 0.5$.

\subsection{Loss Function}
Some methods~\cite{carion2020end,sun2021sparse} minimize the time-consuming post-processing operation by the bipartite matching. Therefore, we extend the bipartite matching from 2D to 3D. Given the ground truth set of objects $y=\left\{y_{i}\right\}_{i=1}^{M}$ and the set of $\mathrm{N}$ prediction $\hat{y}=\left\{\hat{y}_{i}\right\}_{i=1}^{N}$. The matching cost is defined as follows:
\begin{equation}
\mathcal{C}_{\mathrm{match}}=\lambda_{cls}\cdot \mathcal{L}_{cls}+\lambda_{reg}\cdot \mathcal{L}_{reg}+\lambda_{IoU}\cdot \mathcal{L}_{BEV\_IoU}
\label{equation_cost}
\end{equation}
\begin{equation}
\mathcal{C}=\mathop{\arg\min}\limits_{i\in \mathrm{M},j\in \mathrm{N}}\mathcal{C}_{\mathrm{match}}(\hat{y}_{i},y_{j} ) ,
\end{equation}
where, $\lambda_{cls}$, $\lambda_{reg}$, and $\lambda_{IoU}$ are coefficients of each component. $\mathcal{L}_{cls}$ is focal loss~\cite{lin2017focal} of predicted classifications and ground truth category labels. As for regression loss, we adopt the $\ell_{1}$ and BEV $\mathrm{IoU}$ loss $\mathcal{L}_{BEV\_IoU}$ following~\cite{carion2020end,sun2021sparse}. $\mathcal{L}_{reg}$ is the $\ell_{1}$ loss between the normalized predicted boxes and ground truth boxes following~\cite{zhou2018voxelnet}.

The training loss consists of classification, regression, and IoU, applied only to the matched pairs. The IoU loss adopts the rotated 3D DIoU loss~\cite{zhou2019iou} denoted as $\mathcal{L}_{DIoU}$.
\begin{equation}
\mathcal{L}=\lambda_{cls}\cdot \mathcal{L}_{cls}+\lambda_{reg}\cdot \mathcal{L}_{reg}+\lambda_{IoU}\cdot \mathcal{L}_{DIoU} ,
\end{equation}
where the $\lambda_{cls}$, $\lambda_{reg}$, and $\lambda_{IoU}$ represent the weight of corresponding loss, which is the same as the parameters in Eq.~\ref{equation_cost}. We set $\lambda_{cls} = 2$, $\lambda_{reg} = 5$, and $\lambda_{IoU} = 2$.

\subsection{Inference Phase}
The inference procedure of Diff3Det is a denoising process from noise to object boxes. As shown in Fig.~\ref{fig:generater}(b), Diff3Det progressively refines its predictions from boxes sampled in Gaussian distribution. In each sampling step, the random boxes or the estimated boxes from the last step are fed to the decoder to predict the results in the current stage. The proposal boxes for the next step can be computed by the formula~\cite{song2020denoising}:
\begin{equation}
 \boldsymbol{x}_{t-s}=\sqrt{\alpha_{t-s}}\left(\frac{\boldsymbol{x}_{t}-\sqrt{1-\alpha_{t}} \epsilon_{\theta}^{(t)}\left(\boldsymbol{x}_{t}\right)}{\sqrt{\alpha_{t}}}\right)
+\sqrt{1-\alpha_{t-s}-\sigma_{t}^{2}} \cdot \varepsilon_{\theta}^{(t)}\left(\boldsymbol{x}_{t}\right) 
+\sigma_{t} \varepsilon_{t},
\end{equation} 
\begin{equation}
\sigma_{t}=\sqrt{\frac{1 - \alpha_{t} / \alpha_{t-s}}{(1 - \alpha_{t-s}) / (1 - \alpha_{t})}},
\end{equation} 
where $\boldsymbol{x}_{t},\boldsymbol{x}_{t-s}$ represent the proposal boxes in two adjacent steps, $\varepsilon_{\theta}^{(t)}\left(\boldsymbol{x}_{t}\right)$ is the predicted offsets by the decoder, and $\varepsilon_{t}$ is the Gaussian noises. The number of sampling steps is allowed to be equal to or higher than $1$, and the $s$ is the starting time level (i.e., 1000) divided by sampling steps. Besides, the multiple iterations will lead to redundant boxes requiring an added NMS to filter them.

\begin{table*}[t]
    \small
	\setlength{\tabcolsep}{2mm}
	\centering
	\caption{3D object detection results are evaluated on the KITTI validation set with AP calculated by 11 recall positions. We report the average precision of 3D boxes ($AP_{3D}$) and bird’s eye view ($AP_{BEV}$ ) for the car category.}
    \label{tab: results}
		\begin{tabular}{lcccccccc}
			\toprule
			\multirow{2.3}{*}{Method}&\multirow{2.3}{*}{Modality}&\multicolumn{3}{c}{$AP_{3D}$  ($IoU=0.7$)}&\multicolumn{3}{c}{$AP_{BEV}$ ($IoU=0.7$)}\\
			\cmidrule(r){3-5} \cmidrule{6-8}
			&&Easy & Mod. &Hard &Easy & Mod. &Hard     \\
			\cmidrule(r){1-2} \cmidrule(r){3-5} \cmidrule{6-8}
			MV3D~\cite{chen2017multi}  &RGB + LiDAR& 71.29&62.68 &56.56 & 86.55& 78.10  & 76.67  \\
			ContFuse~\cite{liang2018deep}  &RGB + LiDAR&82.54&66.22 &64.04  & 88.81& 85.83 & 77.33   \\
			AVOD-FPN~\cite{ku2018joint}  &RGB + LiDAR& 84.40&74.44 &68.65 & -  &- & -  \\
			F-PointNet~\cite{qi2018frustum}  &RGB + LiDAR& 83.76&70.92 &63.65 & 88.16& 84.02  &76.44   \\
			
			\cmidrule(r){1-2} \cmidrule(r){3-5} \cmidrule{6-8}
		  PointPillars~\cite{lang2019pointpillars} &LiDAR only&79.76&77.01 &74.77  &- &- &-  \\
			VoxelNet~\cite{zhou2018voxelnet}  &LiDAR only&81.97&65.46 &62.85  &89.60&84.81 &78.57   \\
			SECOND~\cite{yan2018second} &LiDAR only &87.43&76.48 &69.10  & \textbf{89.95}&87.07 &79.66 \\
            TANet~\cite{liu2020tanet} &LiDAR only &\textbf{88.21}&77.85 &75.62  &-&-&-\\
            \cmidrule(r){1-2} \cmidrule(r){3-5} \cmidrule{6-8}
			\textbf{Ours (step = 1)}  &LiDAR only& 87.84& \textbf{77.90} & 76.07  & 89.81& 88.24  & 86.68   \\
            \textbf{Ours (step = 4)}  &LiDAR only& 87.38& 77.71 & \textbf{76.44}  & 89.87& \textbf{88.31}  & \textbf{87.05}   \\
            
			\bottomrule
		\end{tabular}

\end{table*} 

\section{Results and Analysis}

\subsection{Dataset}
Our experiments are conducted on the KITTI dataset~\cite{geiger2012we}, which is split into 3717 training and 3769 validation samples. We use the average precision (AP) metric, where the IoU threshold is set to 0.7 for the car category. All experiments are conducted on the car category with easy, moderate, and hard three levels.


\subsection{Implementation Details}
The voxelization range is set to $[0m, 70.4m]$ for $X$ axis, $[-40m, 40m]$ for $Y$ axis, and $[-3m, 1m]$ for $Z$ axis. The voxel size is set to $(0.05m, 0.05m, 0.1m)$. We adopt standard data augmentation techniques~\cite{yan2018second,lang2019pointpillars,shi2019pointrcnn,shi2020points}, including GT sampling, flipping, rotation, scaling, and more. The Diff3Det is trained on 2 NVIDIA RTX 3090 GPUs with batch size 32. We adopt the AdamW~\cite{loshchilov2017decoupled} optimizer with a one-cycle learning rate policy.

\subsection{Main Results}
\begin{figure}[t]
	\begin{center}
		\includegraphics[width=1.0\linewidth]{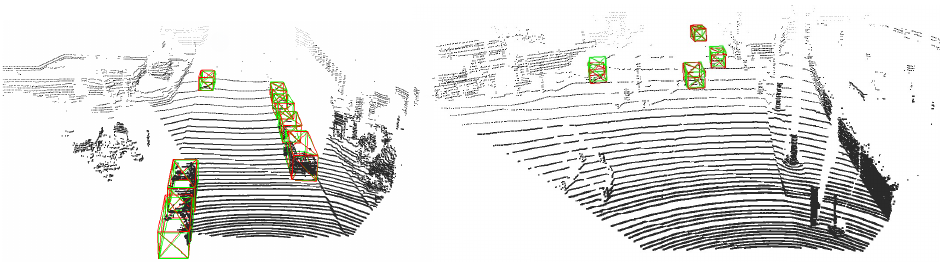}
	\end{center}
        \vspace{-10pt}
	\caption{Qualitative results of Diff3Det on the KITTI validation set. We show the prediction boxes (red) and ground truth boxes (green).}
	\label{fig:vis }
\end{figure}

The main results are shown in Tab.~\ref{tab: results}, where we compare the proposed Diff3Det with classic methods. Our approach achieves better performance compared with the representative anchor-based methods~\cite{zhou2018voxelnet,yan2018second,lang2019pointpillars}. Specifically, one-step Diff3Det outperforms SECOND~\cite{yan2018second} by $1.42\%$ and $6.97\%$ on the moderate and hard levels. Besides, our method exceeds PointPillars~\cite{lang2019pointpillars} $0.89\%$ on the moderate level, $1.3\%$ on the hard level, and $8.08\%$ on the easy level. Qualitative results of one-step Diff3Det are shown in Fig.~\ref{fig:vis }.

When using the multi-step sampling approach commonly used in Diffusion models (i.e., step = 4), the performance improvement is mainly in the hard level of $AP_{3D}$ ($0.37\%$). We argue the main reason is that with the increase in sampling steps, the decoder generates more detection boxes, which is beneficial for detecting difficult samples. However, the large number of boxes may confuse post-processing because of similar predicted classification scores, which causes slight performance damage. The influence of sampling steps will be discussed in the next section.

\subsection{Ablation Studies}
The diffusion-guided proposal generator is the key to Diff3Det. This section explores how it affects performance with extensive ablation studies.

\begin{table}[t]
\small
\setlength{\tabcolsep}{1.25mm}
\centering
\caption{Ablation study of each component in Diff3Det. We gradually add the designed proposal refinements methods by setting random boxes during training as our baseline.}
\label{tab:add_moudel}
\begin{tabular}{ lccccc }
\toprule
Component &Easy&Mod.&Hard&mAP\\
\cmidrule(r){1-1} \cmidrule(r){2-5}
Baseline&82.62&74.21&73.56&76.80\\
+ Corrupted proposals from GT &84.32~(\textcolor{up}{+1.70})&76.17~(\textcolor{up}{+1.96})&74.41~(\textcolor{up}{+0.85})&78.30~(\textcolor{up}{+1.50})\\
+ Resample &85.31~(\textcolor{up}{+0.99})&76.31~(\textcolor{up}{+0.14})&74.48~(\textcolor{up}{+0.08})&78.70~(\textcolor{up}{+0.40})\\
+ Size correlation &86.14~(\textcolor{up}{+0.83})&76.80~(\textcolor{up}{+0.49})&75.29~(\textcolor{up}{+0.81})&79.41~(\textcolor{up}{+0.71})\\
+ Dynamic time step &\textbf{87.84}~(\textcolor{up}{+1.70})&\textbf{77.90}~(\textcolor{up}{+1.10})&\textbf{76.07}~(\textcolor{up}{+0.78})&\textbf{80.61}~(\textcolor{up}{+1.20})\\
\bottomrule
\end{tabular}
\end{table}


\subsubsection{Proposed components.}

To illustrate the effectiveness of the proposed components in the diffusion-guided proposal generator, we conduct ablation studies on our Diff3Det as shown in Tab.~\ref{tab:add_moudel}. Here, our baseline is Diff3Det directly using proposals sampled from a Gaussian distribution for training and inference. When adding boxes corrupted from the ground truth during training, there is a performance gain with mAP of $1.5\%$ over the baseline. Then, we observe that some of the boxes may not have point clouds when selecting initial proposals. Thus, we propose a resampling procedure to ensure each proposal box contains at least several points~(\textit{e.g.,} 5 points). This resampling operation further brings an improvement of $0.99\%$ on the easy level.  Besides, we adopt a size correlation strategy to control the aspect of the size of 3D boxes, which is beneficial to capturing more effective 3D objects. This strategy also brings a performance improvement of $0.71\%$ mAP, which demonstrates the importance of proposal quality. Finally, different from using the fixed time step in most diffusion models, we propose a new dynamic time step to make the whole learning process easier, which produces superior performance with mAP of $80.61\%$ for all three levels. 

\subsubsection{Sampling steps.}

\vspace{5pt}
\begin{wraptable}[11]{r}{0.465\textwidth}
\small
\setlength{\tabcolsep}{2mm}
\centering
\vspace{-0.8cm}
\caption{Effect of sampling steps in test.}
\label{tab: sampling_steps}
\begin{tabular}{ ccccc }
\toprule
 {\multirow{2.3}{*}{Steps}}&\multicolumn{4}{c}{AP$_{40}$(IoU$=0.7$)} \\
\cmidrule(r){2-5}
 & Easy & Mod.& Hard & mAP \\
 \cmidrule(r){1-5}
1  &89.29&79.91 &75.48 &81.56 \\
2  &\textbf{89.56}&80.26 &76.35 &82.06 \\
4  &89.45&\textbf{80.86} &\textbf{77.41} &\textbf{82.57} \\
6  &89.43&80.28 &76.76 &82.16 \\
8  &88.76&80.24 &77.24 &82.08 \\
\bottomrule
\end{tabular}
\end{wraptable}
Diffusion models~\cite{ho2020denoising,song2020denoising} often employ iterative inference to obtain the target distribution. Therefore, we also utilize multiple sampling steps in the test. As shown in Tab.~\ref{tab: sampling_steps}, the average precision (AP) calculated by 40 recall positions exhibits varying degrees of improvement. Notably, the performance at the moderate level is enhanced by $0.95\%$, while the hard level shows an improvement of $1.93\%$ when comparing sampling step=4 to step=0. However, we find that some metrics decreased with calculated by 11 recall positions in Tab.~\ref{tab: results}. We believe it is due to the fact that recall 40 metrics are more accurate~\cite{simonelli2019disentangling} and provide a better reflection of the effectiveness of the iterative approach. We want to assure readers that the slight performance drop in the recall 11 metrics should not lead to any misunderstanding.

\subsubsection{Hyperparameters.}
We perform ablation studies for several important sets of hyperparameters, as shown in Tab.~\ref{tab:parameters}. For the signal scale, which is used to control the signal-to-noise ratio (SNR) of the diffusion process, we empirically find that the value set to 2.0 yields the highest average precision (AP) performance (Tab.~\ref{tab:scale}(a)). For the proposal number $N$, the performance achieves best when its value is set to 300 (Tab.~\ref{tab:number}(b)). The parameter $\eta$ controls the minimum number of point clouds for each proposal, which can effectively improve the quality of proposals. We can observe that the best result is achieved when its value is set to 5 (Tab.~\ref{tab:eta}(c)).

\begin{table}[t]
\small
\caption{Ablation study of hyperparameters.}
\label{tab:parameters}
\centering
\setlength{\tabcolsep}{1mm}
\resizebox{1\linewidth}{!}{
\subtable[Ablation on scale]{
\begin{tabular}{ lccccc }
\toprule
Scale &Easy&Mod.&Hard&mAP\\
\cmidrule(r){1-1} \cmidrule(r){2-5}
0.1 &15.36&14.66&14.35&14.79\\
1.0 &83.87&74.50&73.44&77.27\\
2.0 &\textbf{87.84}&\textbf{77.90}&\textbf{76.07}&\textbf{80.61}\\
\bottomrule
\end{tabular}}
\label{tab:scale}

\subtable[Ablation on proposal number]{
\begin{tabular}{ lccccc }
\toprule
$N$ &Easy&Mod.&Hard&mAP\\
\cmidrule(r){1-1} \cmidrule(r){2-5}
100 &86.59&77.09&75.75&79.81\\
300 &\textbf{87.84}&\textbf{77.90}&\textbf{76.07}&\textbf{80.61}\\
500 &86.25&76.91&75.53&79.63\\
\bottomrule
\end{tabular}}
\label{tab:number}

\subtable[Ablation on $\eta$ in Resample]{
\begin{tabular}{ lccccc }
\toprule
$\eta$ &Easy&Mod.&Hard&mAP\\
\cmidrule(r){1-1} \cmidrule(r){2-5} 
1 &86.44 &77.22 &75.78 &79.81\\
5 &\textbf{87.84} &\textbf{77.90} &\textbf{76.07} &\textbf{80.61} \\
10 &86.69 &77.38 &76.05 &80.04\\
\bottomrule
\end{tabular}}
\label{tab:eta}
}
\vspace{-10pt}
\end{table}

\subsection{Limitation}

The primary limitation is that the proposed method poses difficulty for the decoder to regress prediction from random boxes, leading to a relatively slow convergence speed. Besides, there remains scope for improving the performance of our approach. In the future, we would like to explore fast converging diffusion-based 3D object detection.

\section{Conclusion}
In this paper, we propose a generative-based 3D object detection framework, Diff3Det, by viewing 3D object detection as a denoising diffusion process from noisy boxes to object boxes. Our key idea is to utilize the diffusion method to avoid the empirical heuristics setting of anchors. We hope that our method will provide new insights into the application of generative methods on 3D vision tasks.

\subsubsection{Acknowledgement.}This work was supported by the National Science Fund for Distinguished Young Scholars of China (Grant No.62225603) and the National Undergraduate Training Projects for Innovation and Entrepreneurship (202310487020).

{\small
\bibliographystyle{splncs04}
\bibliography{reference}

\begin{thebibliography}{10}
\providecommand{\url}[1]{\texttt{#1}}
\providecommand{\urlprefix}{URL }
\providecommand{\doi}[1]{https://doi.org/#1}

\bibitem{amit2021segdiff}
Amit, T., Nachmani, E., Shaharbany, T., Wolf, L.: Segdiff: Image segmentation
  with diffusion probabilistic models. arXiv preprint arXiv:2112.00390  (2021)

\bibitem{bai2022transfusion}
Bai, X., Hu, Z., Zhu, X., Huang, Q., Chen, Y., Fu, H., Tai, C.L.: Transfusion:
  Robust lidar-camera fusion for 3d object detection with transformers. In:
  CVPR (2022)

\bibitem{brempong2022denoising}
Brempong, E.A., Kornblith, S., Chen, T., Parmar, N., Minderer, M., Norouzi, M.:
  Denoising pretraining for semantic segmentation. In: CVPR (2022)

\bibitem{carion2020end}
Carion, N., Massa, F., Synnaeve, G., Usunier, N., Kirillov, A., Zagoruyko, S.:
  End-to-end object detection with transformers. In: ECCV (2020)

\bibitem{chen2022diffusiondet}
Chen, S., Sun, P., Song, Y., Luo, P.: Diffusiondet: Diffusion model for object
  detection. arXiv preprint arXiv:2211.09788  (2022)

\bibitem{chen2022generalist}
Chen, T., Li, L., Saxena, S., Hinton, G., Fleet, D.J.: A generalist framework
  for panoptic segmentation of images and videos. arXiv preprint
  arXiv:2210.06366  (2022)

\bibitem{chen2022analog}
Chen, T., Zhang, R., Hinton, G.: Analog bits: Generating discrete data using
  diffusion models with self-conditioning. arXiv preprint arXiv:2208.04202
  (2022)

\bibitem{chen2017multi}
Chen, X., Ma, H., Wan, J., Li, B., Xia, T.: Multi-view 3d object detection
  network for autonomous driving. In: CVPR (2017)

\bibitem{chen2023voxelnext}
Chen, Y., Liu, J., Zhang, X., Qi, X., Jia, J.: Voxelnext: Fully sparse voxelnet
  for 3d object detection and tracking. In: CVPR (2023)

\bibitem{deng2021voxel}
Deng, J., Shi, S., Li, P., Zhou, W., Zhang, Y., Li, H.: Voxel r-cnn: Towards
  high performance voxel-based 3d object detection. In: AAAI (2021)

\bibitem{duan2023diffusiondepth}
Duan, Y., Guo, X., Zhu, Z.: Diffusiondepth: Diffusion denoising approach for
  monocular depth estimation. arXiv preprint arXiv:2303.05021  (2023)

\bibitem{geiger2012we}
Geiger, A., Lenz, P., Urtasun, R.: Are we ready for autonomous driving? the
  kitti vision benchmark suite. In: CVPR (2012)

\bibitem{gu2022vector}
Gu, S., Chen, D., Bao, J., Wen, F., Zhang, B., Chen, D., Yuan, L., Guo, B.:
  Vector quantized diffusion model for text-to-image synthesis. In: CVPR (2022)

\bibitem{sparsedet}
Han, J., Wan, Z., Liu, Z., Feng, J., Zhou, B.: Sparsedet: Towards end-to-end 3d
  object detection. In: VISAPP (2022)

\bibitem{ho2020denoising}
Ho, J., Jain, A., Abbeel, P.: Denoising diffusion probabilistic models. NeurIPS
   (2020)

\bibitem{huang2020epnet}
Huang, T., Liu, Z., Chen, X., Bai, X.: Epnet: Enhancing point features with
  image semantics for 3d object detection. In: ECCV (2020)

\bibitem{ji2023ddp}
Ji, Y., Chen, Z., Xie, E., Hong, L., Liu, X., Liu, Z., Lu, T., Li, Z., Luo, P.:
  Ddp: Diffusion model for dense visual prediction. arXiv preprint
  arXiv:2303.17559  (2023)

\bibitem{ku2018joint}
Ku, J., Mozifian, M., Lee, J., Harakeh, A., Waslander, S.L.: Joint 3d proposal
  generation and object detection from view aggregation. In: IROS (2018)

\bibitem{lang2019pointpillars}
Lang, A.H., Vora, S., Caesar, H., Zhou, L., Yang, J., Beijbom, O.:
  Pointpillars: Fast encoders for object detection from point clouds. In: CVPR
  (2019)

\bibitem{li2023dds3d}
Li, J., Liu, Z., Hou, J., Liang, D.: Dds3d: Dense pseudo-labels with dynamic
  threshold for semi-supervised 3d object detection. In: ICRA (2023)

\bibitem{li2023logonet}
Li, X., Ma, T., Hou, Y., Shi, B., Yang, Y., Liu, Y., Wu, X., Chen, Q., Li, Y.,
  Qiao, Y., et~al.: Logonet: Towards accurate 3d object detection with
  local-to-global cross-modal fusion. In: CVPR (2023)

\bibitem{li2023bevdepth}
Li, Y., Ge, Z., Yu, G., Yang, J., Wang, Z., Shi, Y., Sun, J., Li, Z.: Bevdepth:
  Acquisition of reliable depth for multi-view 3d object detection. In: AAAI
  (2023)

\bibitem{liang2018deep}
Liang, M., Yang, B., Wang, S., Urtasun, R.: Deep continuous fusion for
  multi-sensor 3d object detection. In: ECCV (2018)

\bibitem{lin2017focal}
Lin, T.Y., Goyal, P., Girshick, R., He, K., Doll{\'a}r, P.: Focal loss for
  dense object detection. In: ICCV (2017)

\bibitem{liu2022compositional}
Liu, N., Li, S., Du, Y., Torralba, A., Tenenbaum, J.B.: Compositional visual
  generation with composable diffusion models. In: ECCV (2022)

\bibitem{liu2022petr}
Liu, Y., Wang, T., Zhang, X., Sun, J.: Petr: Position embedding transformation
  for multi-view 3d object detection. In: ECCV (2022)

\bibitem{liu2022epnet++}
Liu, Z., Huang, T., Li, B., Chen, X., Wang, X., Bai, X.: Epnet++: Cascade
  bi-directional fusion for multi-modal 3d object detection. IEEE Transactions
  on Pattern Analysis and Machine Intelligence  (2022)

\bibitem{liu2020tanet}
Liu, Z., Zhao, X., Huang, T., Hu, R., Zhou, Y., Bai, X.: Tanet: Robust 3d
  object detection from point clouds with triple attention. In: AAAI (2020)

\bibitem{loshchilov2017decoupled}
Loshchilov, I., Hutter, F.: Decoupled weight decay regularization. arXiv
  preprint arXiv:1711.05101  (2017)

\bibitem{noh2021hvpr}
Noh, J., Lee, S., Ham, B.: Hvpr: Hybrid voxel-point representation for
  single-stage 3d object detection. In: CVPR (2021)

\bibitem{qi2018frustum}
Qi, C.R., Liu, W., Wu, C., Su, H., Guibas, L.J.: Frustum pointnets for 3d
  object detection from rgb-d data. In: CVPR (2018)

\bibitem{qi2017pointnet++}
Qi, C.R., Yi, L., Su, H., Guibas, L.J.: Pointnet++: Deep hierarchical feature
  learning on point sets in a metric space. In: NeurIPS (2017)

\bibitem{rombach2022high}
Rombach, R., Blattmann, A., Lorenz, D., Esser, P., Ommer, B.: High-resolution
  image synthesis with latent diffusion models. In: CVPR (2022)

\bibitem{shi2022pillarnet}
Shi, G., Li, R., Ma, C.: Pillarnet: High-performance pillar-based 3d object
  detection. In: ECCV (2022)

\bibitem{shi2020pv}
Shi, S., Guo, C., Jiang, L., Wang, Z., Shi, J., Wang, X., Li, H.: Pv-rcnn:
  Point-voxel feature set abstraction for 3d object detection. In: CVPR (2020)

\bibitem{shi2019pointrcnn}
Shi, S., Wang, X., Li, H.: Pointrcnn: 3d object proposal generation and
  detection from point cloud. In: CVPR (2019)

\bibitem{shi2020points}
Shi, S., Wang, Z., Shi, J., Wang, X., Li, H.: From points to parts: 3d object
  detection from point cloud with part-aware and part-aggregation network. IEEE
  Transactions on Pattern Analysis and Machine Intelligence  (2020)

\bibitem{simonelli2019disentangling}
Simonelli, A., Bulo, S.R., Porzi, L., L{\'o}pez-Antequera, M., Kontschieder,
  P.: Disentangling monocular 3d object detection. In: ICCV (2019)

\bibitem{song2020denoising}
Song, J., Meng, C., Ermon, S.: Denoising diffusion implicit models. ICLR
  (2021)

\bibitem{sun2021sparse}
Sun, P., Zhang, R., Jiang, Y., Kong, T., Xu, C., Zhan, W., Tomizuka, M., Li,
  L., Yuan, Z., Wang, C., et~al.: Sparse r-cnn: End-to-end object detection
  with learnable proposals. In: CVPR (2021)

\bibitem{xiong2023cape}
Xiong, K., Gong, S., Ye, X., Tan, X., Wan, J., Ding, E., Wang, J., Bai, X.:
  Cape: Camera view position embedding for multi-view 3d object detection. In:
  CVPR (2023)

\bibitem{yan2018second}
Yan, Y., Mao, Y., Li, B.: Second: Sparsely embedded convolutional detection.
  Sensors  (2018)

\bibitem{yang20203dssd}
Yang, Z., Sun, Y., Liu, S., Jia, J.: 3dssd: Point-based 3d single stage object
  detector. In: CVPR (2020)

\bibitem{yin2021center}
Yin, T., Zhou, X., Krahenbuhl, P.: Center-based 3d object detection and
  tracking. In: CVPR (2021)

\bibitem{zhang2023sam3d}
Zhang, D., Liang, D., Yang, H., Zou, Z., Ye, X., Liu, Z., Bai, X.: Sam3d:
  Zero-shot 3d object detection via segment anything model. arXiv preprint
  arXiv:2306.02245  (2023)

\bibitem{2023asimple}
Zhang, D., Liang, D., Zou, Z., Li, J., Ye, X., Liu, Z., Tan, X., Bai, X.: A
  simple vision transformer forweakly semi-supervised 3d object detection. In:
  ICCV (2023)

\bibitem{zhang2022not}
Zhang, Y., Hu, Q., Xu, G., Ma, Y., Wan, J., Guo, Y.: Not all points are equal:
  Learning highly efficient point-based detectors for 3d lidar point clouds.
  In: CVPR (2022)

\bibitem{zhou2019iou}
Zhou, D., Fang, J., Song, X., Guan, C., Yin, J., Dai, Y., Yang, R.: Iou loss
  for 2d/3d object detection. In: 3DV (2019)

\bibitem{zhou2018voxelnet}
Zhou, Y., Tuzel, O.: Voxelnet: End-to-end learning for point cloud based 3d
  object detection. In: CVPR (2018)

\end{thebibliography}
}

\end{document}